\documentclass{research}

\definecolor{basegray}{RGB}{245,245,245}

\title{Multiple Turns, Short Horizons: Measuring Effective-Horizon Compression in Jailbreaks}

\author[1,2]{Zhipeng Wei}
\author[1,3]{Michael Lee}
\author[1,4]{N. Benjamin Erichson}

\affiliation[1]{International Computer Science Institute}
\affiliation[2]{UC Berkeley}
\affiliation[3]{DSO National Laboratories}
\affiliation[4]{Lawrence Berkeley National Lab}

\correspondence{Zhipeng Wei @ \email{zwei@icsi.berkeley.edu} and N. Benjamin Erichson @ \email{erichson@icsi.berkeley.edu}.}

\abstract{Multi-turn jailbreaks use feedback from a victim large language model to adapt their queries over several turns. However, once an effective trajectory has been found, it is unclear how much of the earlier query history is still needed. In this work, we study this by removing early attacker queries and running the remaining queries again in a new conversation while regenerating the victim's responses. We define the \emph{effective horizon} as the shortest remaining query sequence that recovers the same attack progress. Empirically, we find that roughly half of the original query sequence is sufficient on average to recover the same attack progress. We call this phenomenon \emph{effective-horizon compression}. We show that this shorter history can recover the best point reached during an unsuccessful attack and provide a better starting point for continuing the attack. It also helps identify which earlier queries should receive less learning signal during attacker training, and downweighting those queries improves performance. We further find that short reward-propagation windows preserve calibrated defense performance close to full-trajectory returns. These results show that multi-turn attacks can contain more queries than are needed to recover the progress reached later in the interaction.}

\begin{document}

\maketitle

\section{Introduction}
\label{sec:introduction}

Large language models (LLMs) are vulnerable to jailbreak attacks that can make them produce harmful responses despite safety alignment~\citep{zou2023universal, shen2024anything, chao2025jailbreaking}. Multi-turn jailbreaks are especially effective because each turn gives the attacker information that can steer the queries that follow~\citep{russinovich2025great, yang2024chain, ren2025llms, rahman2025x}. However, this adaptivity can make the attack slow and costly, since the attacker may need several rounds of interaction before finding a query sequence that works.

Multi-turn attacks can be viewed as an adaptive search over possible attack trajectories. The attacker uses feedback from each interaction to decide how to proceed, often taking several steps before reaching an effective query sequence. However, once such a trajectory has been found, some of its earlier queries may no longer be needed. A shorter sequence of the later queries may be enough to recover the same attack progress. We therefore ask:

\begin{tcolorbox}[
  colback=parchment,
  colframe=sandborder,
  boxrule=0.3pt,
  left=8pt,
  right=8pt,
  top=1pt,
  bottom=1pt
]
\centering
\emph{How much prior query history is needed to recover the same attack progress at a later turn?}
\end{tcolorbox}

\begin{wrapfigure}[19]{r}{0.52\columnwidth}
    \centering
    \vspace{+0.2cm}
    \includegraphics[width=0.5\columnwidth]{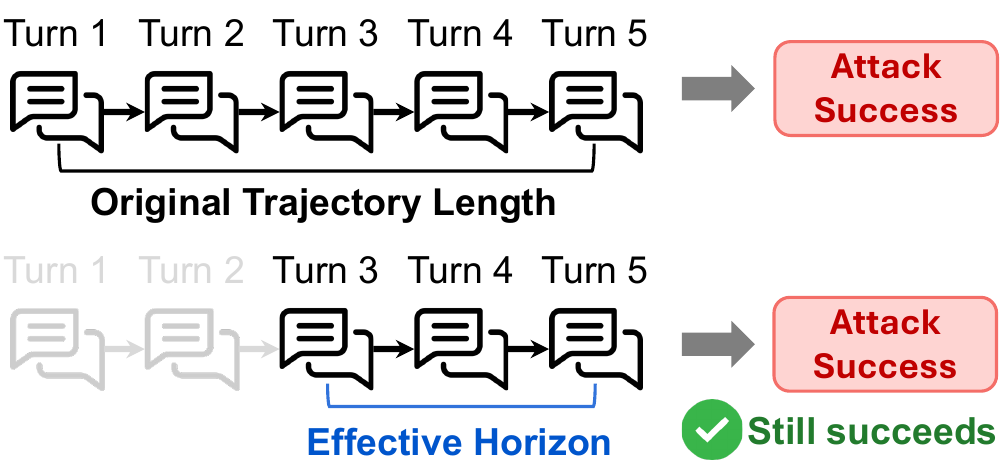}
    \caption{A later point in a multi-turn attack may require less query history than the full trajectory that led to it. We measure this by removing the earliest attacker queries and running the remaining queries again in a new conversation, while the victim generates new responses. The \emph{effective horizon} is the shortest remaining query sequence that recovers the same attack progress reached at the target turn. In this example, the first two queries can be removed and the later queries still recover the same progress.}
    \label{fig:effective_horizon}
    \vspace{-1pt}
\end{wrapfigure}

To study this question, we remove earlier queries from successful attack trajectories and test whether the remaining queries can recover the same attack progress. As illustrated in Figure~\ref{fig:effective_horizon}, we take the last few attacker queries from an existing trajectory and run them again in a new conversation. The attacker queries are kept exactly the same, while the victim generates new responses to them. If this shorter sequence reaches the same level of attack progress as the original trajectory, within a small tolerance, then the removed queries were not needed for the retained queries to remain effective. We define the \emph{effective horizon} as the length of the shortest sequence that satisfies this condition.

We find that the effective horizon is often much shorter than the original trajectory generated by either X-Teaming~\citep{rahman2025x} or TROJail~\citep{xiong2026trojail}. Across our evaluated settings, roughly half of the original query sequence is sufficient on average to recover the same attack progress. We refer to this reduction as \emph{effective-horizon compression}. An attack trajectory can therefore often be shortened substantially while remaining similarly effective.
This suggests that the full sequence of queries used to produce an attack may contain more history than is ultimately needed for the attack to remain effective.

This differs from shortening an attack by rewriting it. M2S~\citep{ha-etal-2025-one}, for example, uses a multi-turn dialogue to construct a new single-turn jailbreak prompt. Such a prompt can incorporate information from the entire dialogue. We leave the original attacker queries unchanged and only remove queries that came before them. The effective horizon therefore measures how much of the original query sequence is still needed to recover the same attack progress.

Effective-horizon compression also gives us a way to make better use of unsuccessful attacks. A failed trajectory may have made substantial progress at an earlier turn before later queries moved the attack in a worse direction. Rather than continuing from the end of such a trajectory, we return to its highest-scoring turn. We then use the effective horizon to keep only the shortest query history needed to recover the progress reached at that turn, and continue the attack from there. We call this \emph{Effective-Horizon Context Recovery} (EH-CR). Across X-Teaming~\citep{rahman2025x} and ActorBreaker~\citep{ren2025llms}, EH-CR improves attack continuation compared with both keeping the complete failed trajectory and returning to the highest-scoring turn with its full history.

Effective-horizon compression also suggests a different way to train multi-turn attackers. When an attack succeeds, the training algorithm must decide which of the queries in the trajectory should be reinforced for producing that outcome. TROJail~\citep{xiong2026trojail} uses the final attack outcome as a learning signal for every attacker query in the trajectory. If the same attack progress can be recovered from only the last few queries, however, much earlier queries may not deserve the same learning signal. This does not mean that the earlier queries were useless, since they may have helped the attacker discover what to ask later. It does suggest that their contribution is different. We therefore introduce \emph{Effective-Horizon Credit Assignment} (EH-CA), which keeps the full learning signal for queries within the effective horizon and reduces it for earlier queries. Across nine victim-model and benchmark pairs, this partial downweighting improves ASR@3 by 6.9 points on average over TROJail's original credit assignment.

We ask whether the same idea also matters for defense training. TurnGate~\citep{shen2026one} assigns a reward to each \textsc{Pass} or \textsc{Block} decision, but trains an earlier decision using both its own reward and the rewards obtained at later turns. In the original method, this learning signal extends through the entire remaining conversation. We test whether such long-range reward propagation is actually necessary by limiting each decision to rewards from only a short number of subsequent turns. After calibrating the decision threshold, performance changes very little across these different reward windows. In particular, $\mathrm{F1}_{\phi_2}$ has a standard deviation of only $0.16\%$, and even using only the reward from the current turn performs nearly the same as using rewards from the full remaining trajectory. This suggests that a long interaction does not necessarily require learning signals to propagate across the entire conversation. Together with our attacker-training results, it shows that the useful range of a training signal can be much shorter than the trajectory in which it was generated.

Our contributions are:
\begin{itemize}[leftmargin=*]

    \item \textbf{Later attack progress often requires only part of the query history.}
    We introduce the \emph{effective horizon}, the shortest sequence of later attacker queries that recovers the same attack progress as the original trajectory. Across X-Teaming and TROJail, roughly half of the original query sequence is sufficient on average. We call this \emph{effective-horizon compression} (\cref{sec:effective_horizon}).

    \item \textbf{Shorter histories can provide better starting points for continuing an attack.}
    An unsuccessful attack may reach a promising point and then move away from it. EH-CR returns to the highest-scoring turn and keeps only the query history needed to recover that progress. This improves continuation over the failed trajectory and the full history leading to the same turn (\cref{sec:applications}).

    \item \textbf{Learning signals need not extend across the full interaction.}
    \emph{Effective-Horizon Credit Assignment} (EH-CA) reduces the learning signal given to queries before the effective horizon and improves attacker training across all nine evaluated settings. In defense training, calibrated performance also changes little when rewards are propagated over much shorter ranges. The useful range of a training signal can therefore be shorter than the trajectory that produced it (\cref{sec:credit_assignment}).

\end{itemize}

\section{Related Work}

Our work relates to two lines of research on multi-turn jailbreaks: methods that construct attacks over several turns and defenses that use information across a conversation. We study how much of the earlier query history remains necessary once later attack queries have been produced.

\textbf{Multi-turn Jailbreak Attacks.}
Search-based attacks generate queries during the interaction and adapt them to the victim's responses. Crescendo~\citep{russinovich2025great}, FITD~\citep{weng2025foot}, and CoA~\citep{yang2024chain} progressively develop an attack across turns, while ActorBreaker~\citep{ren2025llms}, X-Teaming~\citep{rahman2025x}, Trojan Knowledge~\citep{wei2025trojan}, and PLAGUE~\citep{bhuiya2025plague} incorporate planning, exploration, or coordination among multiple agents. Training-based methods instead learn policies for generating multi-turn attacks from supervised or reinforcement-learning signals~\citep{zhao2025siren, xiong2026trojail, feng2026sema}. M2S~\citep{ha-etal-2025-one} takes a different approach by rewriting a multi-turn dialogue into a new single-turn jailbreak prompt. Our analysis keeps the original attacker queries unchanged and measures how much of their preceding query history can be removed.

\textbf{Multi-turn Jailbreak Defenses.}
Multi-turn defenses use conversation history to detect harmful intent that may emerge only over several turns. Existing methods reason over the accumulated dialogue~\citep{kuo2025safety}, maintain information across turns using graph or recurrent representations~\citep{huang2025attention, albrethsen2026deepcontext}, or accumulate evidence of risk over time~\citep{ma2026thrd}. D-Judge modifies victim responses to interrupt attack progression~\citep{gong2026d}, while TRACE~\citep{miah2026trace} and TurnGate~\citep{shen2026one} use trajectory-level reasoning or learned policies to decide when to block an interaction. Our defense experiments leave the observed conversation unchanged and instead study how far training rewards need to propagate across turns.

\section{On Query-History Dependence in Multi-Turn Jailbreaks}
\label{sec:effective_horizon}

We aim to understand how much prior query history is needed to recover the attack progress reached at a later turn. We study this by taking an existing attack trajectory and selecting a target turn. We then remove attacker queries from the beginning of the sequence and run the remaining queries again in a new conversation, allowing the victim to generate new responses. If the shorter sequence reaches about the same attack score as the original query sequence, then the removed queries were not needed. The shortest query sequence that satisfies this condition defines the \emph{effective horizon}.

\subsection{Effective Horizon}

We now make precise what we mean by the \emph{effective horizon}. For any point in an attack trajectory, we ask how many of the most recent attacker queries are sufficient to recover the progress reached at that point. Consider a trajectory with attacker query sequence \(\mathbf{q}_{1:T}=(q_1,\ldots,q_T)\), and let \(j\) denote the turn whose attack progress we want to recover. Let \(v_j\) be the attack score reached at that turn. For a candidate length \(k\), we remove the earlier queries and run only the last \(k\) queries, \(\mathbf{q}_{j-k+1:j}\), in a new conversation. The victim generates new responses, producing an attack score \(\widetilde v_{j,k}\).

We say that a shorter sequence recovers the original attack progress if
\(\widetilde v_{j,k}\geq v_j-\epsilon\), where \(\epsilon\geq0\) is the maximum score decrease we allow. The value of \(\epsilon\) depends on the scoring function and is specified for each experiment. We only remove queries from the beginning of the sequence, so each candidate is a contiguous suffix of the original query history. The \emph{effective horizon} is the length of the shortest such suffix that recovers the original progress:
\begin{equation}
H_{\mathrm{eff}}(\mathbf{q}_{1:j})
=
\min\left\{
k\in\{1,\ldots,j\}
\;\middle|\;
\widetilde v_{j,k}\geq v_j-\epsilon
\right\}.
\label{eq:effective_horizon}
\end{equation}

The effective horizon therefore tells us how much recent query history is sufficient to recover the progress reached at turn \(j\). If \(H_{\mathrm{eff}}=2\), for example, the last two attacker queries are sufficient and all earlier queries can be removed. If no tested suffix recovers the original progress, the effective horizon is undefined for that target turn.
This definition isolates dependence on the preceding query history. We do not rewrite the retained queries or remove arbitrary turns; we only remove a prefix and regenerate the victim responses. A short effective horizon therefore means that the later queries can recover the target progress with little of the earlier query history still present. It does not mean that the removed turns were unnecessary for discovering those later queries.
To compare trajectories of different lengths, we use the \emph{effective-horizon ratio}, \(\mathrm{EHR}=H_{\mathrm{eff}}/j\). An EHR of \(1\) means that the full query history is needed under our replay criterion, while smaller values indicate greater compression. We refer to this reduction as \emph{effective-horizon compression}. The EHR therefore lets us compare how much query history remains necessary across attacks of different lengths.

\begin{figure}[t]
    \centering
    \includegraphics[width=0.98\linewidth]{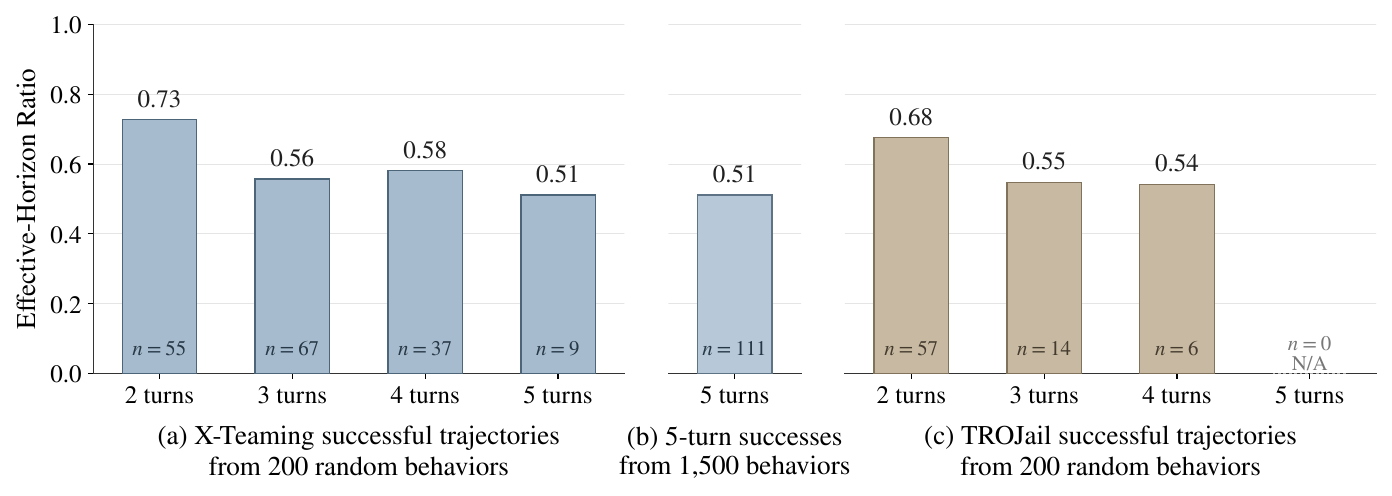}
    \vspace{-0.3cm}
    \caption{The attack progress reached by successful multi-turn attacks can often be recovered from only about half of their original query history. We measure this with the EHR, which divides the shortest query suffix that recovers the original attack progress by the full trajectory length. Across both X-Teaming and TROJail, trajectories of three or more turns have average EHRs close to \(0.5\)--\(0.6\). The larger X-Teaming evaluation in (b) confirms this pattern, with five-turn attacks reaching an average EHR of \(0.512\) over \(111\) successful trajectories. Panels (a) and (c) use 200 WildJailbreak behaviors, while (b) uses 1,500; \(n\) denotes the number of successful trajectories.
    }
    \label{fig:effective_horizon_bar}
\end{figure}

\subsection{Effective-Horizon Compression Across Attack Paradigms}

We next test whether effective-horizon compression appears across different multi-turn attack methods. For successful trajectories from X-Teaming~\citep{rahman2025x} and TROJail~\citep{xiong2026trojail}, we measure the effective horizon of the final turn. Both are evaluated on 200 WildJailbreak behaviors~\citep{jiang2024wildteaming} with a maximum of five turns. 
See experimental details in Appendix~\ref{app:more_details_ehc}.
Figure~\ref{fig:effective_horizon_bar}(a) and (c) show a similar pattern for both attacks. For two-turn trajectories, the average EHR is \(0.73\) for X-Teaming and \(0.68\) for TROJail. Since a two-turn trajectory can only have an EHR of \(0.5\) or \(1\), these values mean that more than half of the successful attacks recover their original progress using only the final query. For longer trajectories, the average EHR remains around \(0.5\)--\(0.6\). This means that, even when an attack unfolds over several turns, only about half of the query history is needed on average to recover the progress reached at the end. The removed turns may still be necessary for current methods to discover the later queries. Their redundancy after the fact nevertheless suggests that there may be shorter paths to the same attack progress, and motivates future methods that
discover these shorter paths more directly. Appendix~\ref{app:case_study} provides a case study.

\begin{figure}[!t]
    \centering
    \includegraphics[width=\linewidth]{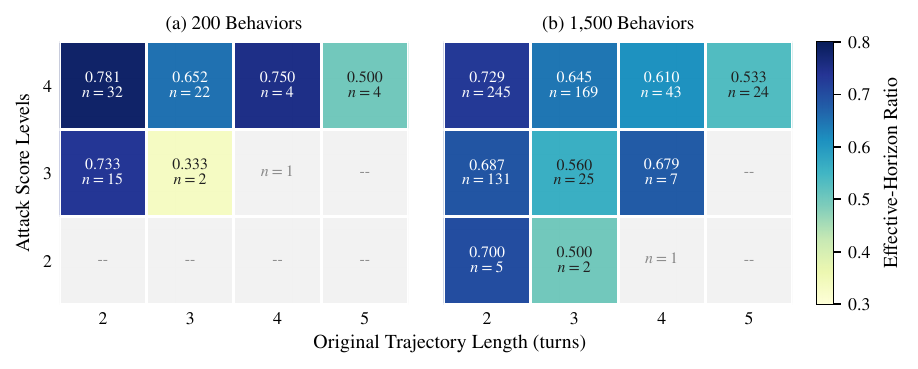}
    \vspace{-0.8cm}
    \caption{
    Effective-horizon compression also appears in unsuccessful attacks. We truncate each X-Teaming trajectory at the first turn that reaches its highest attack score and measure how much query history is needed to recover that progress. Most supported groups have EHR below \(1\), including high-scoring intermediate states. In the larger 1,500-behavior evaluation, score-\(4\) trajectories decrease from an average EHR of \(0.729\) at two turns to \(0.533\) at five turns. This shows that attack progress can be recovered from a shorter history even before the attack succeeds.
    }
    \label{fig:effective_horizon-failed}
\end{figure}

Because the initial sample contains few successful five-turn X-Teaming trajectories, we repeat that setting with 1,500 behaviors. The resulting 111 successful attacks have an average EHR of \(0.51\), closely matching the smaller evaluation in Figure~\ref{fig:effective_horizon_bar}(b). The same effect also appears before an attack succeeds. Figure~\ref{fig:effective_horizon-failed} measures the effective horizon at the highest-scoring point reached by unsuccessful X-Teaming trajectories. Across supported trajectory lengths and score levels, these intermediate states can often be recovered from only part of the preceding query history. In the larger evaluation, for example, trajectories reaching score \(4\) decrease from an average EHR of \(0.729\) at two turns to \(0.533\) at five turns. Effective-horizon compression therefore applies not only to successful attacks, but also to promising intermediate points that an attacker may want to return to. This observation motivates the continuation method in the next section.

\section{Context Recovery in Search-Based Attacks}
\label{sec:applications}

In the previous section, we showed that the progress reached by a multi-turn attack can often be recovered from much less query history than the full trajectory. We now ask whether this shorter history can help continue attacks that have not yet succeeded. A failed trajectory may have reached substantial progress at an earlier turn before later queries moved it in a worse direction. Returning to that earlier point removes those later failed turns, but still keeps all of the query history that led there. If only part of that history is needed to recover the same progress, we can restart the attack from a shorter context. This gives the attacker a better place to continue searching, using its remaining turns from a point that preserves the best progress reached so far without carrying unnecessary history.

\textbf{Effective-Horizon Context Recovery (EH-CR).} Given an unsuccessful trajectory, EH-CR first returns to the turn with the highest attack score,
\begin{equation}
j^\star
=
\arg\max_{1\leq j\leq T} v_j .
\label{eq:ehcr_peak}
\end{equation}
We then apply Equation~\ref{eq:effective_horizon} to the query history up to \(j^\star\) and keep only the shortest suffix that recovers the progress reached there. We run this shorter sequence again in a new conversation and continue the attack from the resulting response. If no shorter suffix recovers the same progress, we keep the full prefix up to \(j^\star\).
EH-CR removes two kinds of unnecessary history. It first discards the turns that came after the highest-scoring point, since those turns moved the attack away from its best progress. It then shortens the history leading up to that point by removing earlier queries that are no longer needed to recover it. The attack therefore restarts from its best recovered point using the shortest history that can reproduce that progress.

\paragraph{Does shorter history improve continuation?} We test this with X-Teaming and ActorBreaker on 159 HarmBench behaviors. Each attack first runs for five turns, and unsuccessful trajectories receive five additional turns. We compare continuing from the full failed trajectory, returning to the highest-scoring turn with its full history (\emph{Peak Context}), and returning to that same turn using only its effective suffix (EH-CR).
See experimental details in Appendix~\ref{app:more_details_cr}. 
Table~\ref{tab:recovery_gain} shows that EH-CR gives the largest improvement for every attack--victim pair. For X-Teaming, it improves ASR by \(17.4\) points on average, compared with \(10.5\) for Peak Context. For ActorBreaker, the gains are \(19.2\) and \(14.2\) points, respectively. The comparison with Peak Context is especially informative because both methods restart from the same point. The only difference is how much earlier history they retain. This isolates the effect of shorter retained history. 
\begin{wrapfigure}[20]{r}{0.58\columnwidth}
    \centering
    \includegraphics[width=0.58\columnwidth]{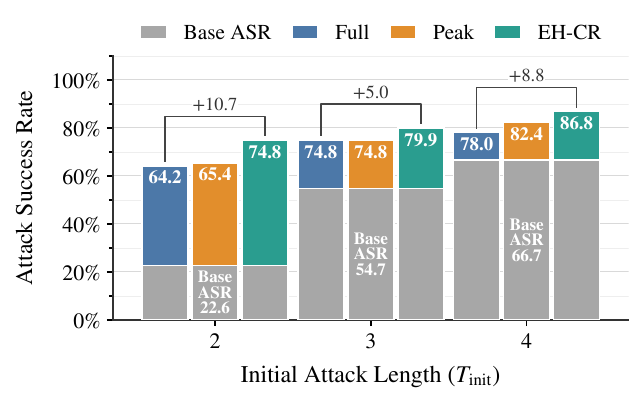}
    \vspace{-0.8cm}
    \caption{
    EH-CR consistently improves attack continuation across initial lengths.
    Each method receives five additional turns after an initial attack of \(T_{\mathrm{init}}\in\{2,3,4\}\). EH-CR consistently achieves the highest final ASR, improving over Full Context by \(8.2\) points on average and showing that recovering a shorter effective history remains useful even as the amount of prior interaction changes.
    }
    \label{fig:initial_length}
\end{wrapfigure}
The advantage of EH-CR therefore shows that history which is unnecessary for recovering attack progress can also hinder attack continuation.

\begin{figure}[!t]
    \centering
    \includegraphics[width=\linewidth]{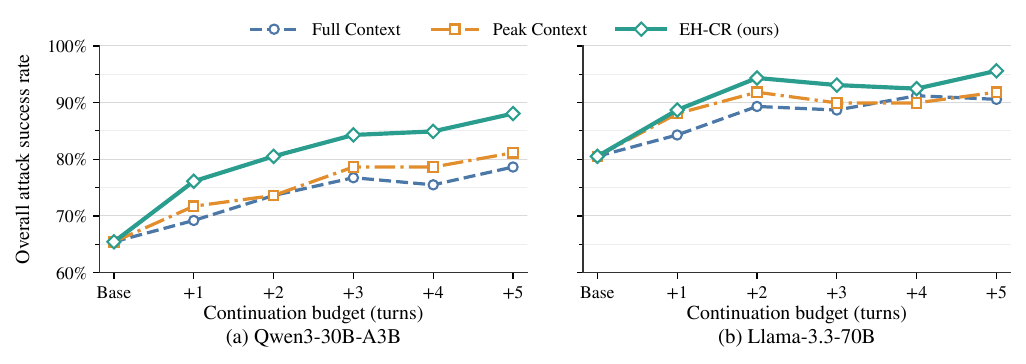}
    \vspace{-0.7cm}
    \caption{
    EH-CR remains more effective across continuation budgets. Starting from 5-turn initial trajectories, we vary 1–5 additional turns on 159 HarmBench behaviors for Qwen3-30B-A3B and Llama-3.3-70B. EH-CR achieves higher final ASR than Full and Peak for both victims at every continuation budget, with the advantage already appearing after a single additional turn.
    }
    \label{fig:effective_horizon-length-changed}
\end{figure}

This pattern also holds when we vary how much interaction occurs before continuation. As shown in Figure~\ref{fig:initial_length}, EH-CR achieves the highest final ASR for initial trajectories of two, three, and four turns, improving over Full Context by \(8.2\) points on average. We also vary the number of turns available for continuation. Figure~\ref{fig:effective_horizon-length-changed} shows that EH-CR already improves final ASR after one or two additional turns, and remains more effective as the continuation budget increases. Together, these results show that the effective horizon is useful beyond measurement. The shorter history identified by replay is not only sufficient to recover earlier attack progress; it can provide a better starting point for continuing the search. EH-CR is intended as a context-selection method rather than a way to reduce query count; query-cost results are reported in Appendix~\ref{app:victim}.

\begin{table*}[!t]
\centering
\caption{
Attack continuation after five-turn initial trajectories. Each unsuccessful attack receives five additional turns. Full continues from the complete trajectory, Peak returns to the highest-scoring turn with its full prefix, and EH-CR returns to the same turn using only its effective suffix. Entries report \(\Delta\mathrm{ASR}\), with final ASR in parentheses. Best values are shown in bold. EH-CR performs best for every attack--victim pair, showing that the shorter history is not only sufficient to recover progress but can provide a better continuation context despite using less of the original history.
}
\label{tab:recovery_gain}

\resizebox{\columnwidth}{!}{
\begin{tabular}{
l
c
c
c
>{\columncolor{basegray}}c
c
c
c
>{\columncolor{basegray}}c
}

\toprule

&
\multicolumn{4}{c}{X-Teaming}
&
\multicolumn{4}{c}{ActorBreaker}
\\

\cmidrule(lr){2-5}
\cmidrule(lr){6-9}

\multirow{2}{*}{Victim Model}
&
\multirow{2}{*}{\color{gray}Base ASR}
&
\multicolumn{3}{c}{$\Delta$ASR from Continuation $\uparrow$}
&
\multirow{2}{*}{\color{gray}Base ASR}
&
\multicolumn{3}{c}{$\Delta$ASR from Continuation $\uparrow$}
\\

\cmidrule(lr){3-5}
\cmidrule(lr){7-9}

&
&
Full
&
Peak
&
EH-CR
&
&
Full
&
Peak
&
EH-CR
\\

\midrule

Qwen3-30B-A3B
&
\color{gray}67.9
&
\begin{tabular}[c]{@{}c@{}}
+13.2\\[-1pt]
{\scriptsize\color{gray}(81.1)}
\end{tabular}
&
\begin{tabular}[c]{@{}c@{}}
+15.7\\[-1pt]
{\scriptsize\color{gray}(83.6)}
\end{tabular}
&
\begin{tabular}[c]{@{}c@{}}
\textbf{+22.6}\\[-1pt]
{\scriptsize\color{gray}(90.5)}
\end{tabular}
&
\color{gray}37.1
&
\begin{tabular}[c]{@{}c@{}}
+20.8\\[-1pt]
{\scriptsize\color{gray}(57.9)}
\end{tabular}
&
\begin{tabular}[c]{@{}c@{}}
+20.1\\[-1pt]
{\scriptsize\color{gray}(57.2)}
\end{tabular}
&
\begin{tabular}[c]{@{}c@{}}
\textbf{+22.0}\\[-1pt]
{\scriptsize\color{gray}(59.1)}
\end{tabular}
\\

Llama-3.3-70B
&
\color{gray}80.5
&
\begin{tabular}[c]{@{}c@{}}
+10.1\\[-1pt]
{\scriptsize\color{gray}(90.6)}
\end{tabular}
&
\begin{tabular}[c]{@{}c@{}}
+11.3\\[-1pt]
{\scriptsize\color{gray}(91.8)}
\end{tabular}
&
\begin{tabular}[c]{@{}c@{}}
\textbf{+15.1}\\[-1pt]
{\scriptsize\color{gray}(95.6)}
\end{tabular}
&
\color{gray}32.7
&
\begin{tabular}[c]{@{}c@{}}
+22.0\\[-1pt]
{\scriptsize\color{gray}(54.7)}
\end{tabular}
&
\begin{tabular}[c]{@{}c@{}}
+23.9\\[-1pt]
{\scriptsize\color{gray}(56.6)}
\end{tabular}
&
\begin{tabular}[c]{@{}c@{}}
\textbf{+27.7}\\[-1pt]
{\scriptsize\color{gray}(60.4)}
\end{tabular}
\\

Muse-Glimmer-30B
&
\color{gray}15.1
&
\begin{tabular}[c]{@{}c@{}}
+3.1\\[-1pt]
{\scriptsize\color{gray}(18.2)}
\end{tabular}
&
\begin{tabular}[c]{@{}c@{}}
+3.1\\[-1pt]
{\scriptsize\color{gray}(18.2)}
\end{tabular}
&
\begin{tabular}[c]{@{}c@{}}
\textbf{+10.7}\\[-1pt]
{\scriptsize\color{gray}(25.8)}
\end{tabular}
&
\color{gray}13.2
&
\begin{tabular}[c]{@{}c@{}}
+10.7\\[-1pt]
{\scriptsize\color{gray}(23.9)}
\end{tabular}
&
\begin{tabular}[c]{@{}c@{}}
+8.8\\[-1pt]
{\scriptsize\color{gray}(22.0)}
\end{tabular}
&
\begin{tabular}[c]{@{}c@{}}
\textbf{+13.8}\\[-1pt]
{\scriptsize\color{gray}(27.0)}
\end{tabular}
\\

DeepSeek-V4.1-Flash
&
\color{gray}23.9
&
\begin{tabular}[c]{@{}c@{}}
+10.1\\[-1pt]
{\scriptsize\color{gray}(34.0)}
\end{tabular}
&
\begin{tabular}[c]{@{}c@{}}
+13.2\\[-1pt]
{\scriptsize\color{gray}(37.1)}
\end{tabular}
&
\begin{tabular}[c]{@{}c@{}}
\textbf{+18.9}\\[-1pt]
{\scriptsize\color{gray}(42.8)}
\end{tabular}
&
\color{gray}11.3
&
\begin{tabular}[c]{@{}c@{}}
+4.4\\[-1pt]
{\scriptsize\color{gray}(15.7)}
\end{tabular}
&
\begin{tabular}[c]{@{}c@{}}
+5.7\\[-1pt]
{\scriptsize\color{gray}(17.0)}
\end{tabular}
&
\begin{tabular}[c]{@{}c@{}}
\textbf{+11.3}\\[-1pt]
{\scriptsize\color{gray}(22.6)}
\end{tabular}
\\

GPT-5.6 Luna
&
\color{gray}13.8
&
\begin{tabular}[c]{@{}c@{}}
+8.2\\[-1pt]
{\scriptsize\color{gray}(22.0)}
\end{tabular}
&
\begin{tabular}[c]{@{}c@{}}
+11.3\\[-1pt]
{\scriptsize\color{gray}(25.1)}
\end{tabular}
&
\begin{tabular}[c]{@{}c@{}}
\textbf{+20.1}\\[-1pt]
{\scriptsize\color{gray}(33.9)}
\end{tabular}
&
\color{gray}8.2
&
\begin{tabular}[c]{@{}c@{}}
+6.9\\[-1pt]
{\scriptsize\color{gray}(15.1)}
\end{tabular}
&
\begin{tabular}[c]{@{}c@{}}
+6.9\\[-1pt]
{\scriptsize\color{gray}(15.1)}
\end{tabular}
&
\begin{tabular}[c]{@{}c@{}}
\textbf{+16.4}\\[-1pt]
{\scriptsize\color{gray}(24.6)}
\end{tabular}
\\

Gemini 3.8 Flash
&
\color{gray}46.5
&
\begin{tabular}[c]{@{}c@{}}
+8.2\\[-1pt]
{\scriptsize\color{gray}(54.7)}
\end{tabular}
&
\begin{tabular}[c]{@{}c@{}}
+8.2\\[-1pt]
{\scriptsize\color{gray}(54.7)}
\end{tabular}
&
\begin{tabular}[c]{@{}c@{}}
\textbf{+17.0}\\[-1pt]
{\scriptsize\color{gray}(63.5)}
\end{tabular}
&
\color{gray}26.4
&
\begin{tabular}[c]{@{}c@{}}
+20.1\\[-1pt]
{\scriptsize\color{gray}(46.5)}
\end{tabular}
&
\begin{tabular}[c]{@{}c@{}}
+19.5\\[-1pt]
{\scriptsize\color{gray}(45.9)}
\end{tabular}
&
\begin{tabular}[c]{@{}c@{}}
\textbf{+23.9}\\[-1pt]
{\scriptsize\color{gray}(50.3)}
\end{tabular}
\\

\midrule

Avg.
&
\color{gray}-
&
\begin{tabular}[c]{@{}c@{}}
+8.8
\end{tabular}
&
\begin{tabular}[c]{@{}c@{}}
+10.5
\end{tabular}
&
\begin{tabular}[c]{@{}c@{}}
\textbf{+17.4}
\end{tabular}
&
\color{gray}-
&
\begin{tabular}[c]{@{}c@{}}
+14.2
\end{tabular}
&
\begin{tabular}[c]{@{}c@{}}
+14.2
\end{tabular}
&
\begin{tabular}[c]{@{}c@{}}
\textbf{+19.2}
\end{tabular}
\\

\bottomrule

\end{tabular}}
\end{table*}

\section{Effective Horizon and Temporal Credit Assignment}
\label{sec:credit_assignment}

The previous sections showed that attack progress can often be recovered from much less query history than the full trajectory, and that this shorter history can provide a better continuation context. We now ask whether the same idea should change how learning signals are assigned during training. When an attack succeeds, a training method must decide how much that final outcome should reinforce each earlier query. If the same progress can be recovered from only the last few queries, giving every turn the same learning signal may be too coarse. Earlier queries may still have helped the attacker discover what to ask later, so we do not assume that their contribution is zero. We therefore study whether they should receive less credit for the final outcome, and ask the related question in defense training of how far turn-level rewards need to influence earlier decisions.

\subsection{Outcome Credit in Attacker Training}
\label{sec:51}

\paragraph{Effective-Horizon Credit Assignment (EH-CA).}
TROJail~\citep{xiong2026trojail} provides a direct way to test this idea because it uses the final attack outcome as a learning signal for every attacker query in the trajectory. An early query therefore receives the same outcome signal as a query near the end of the attack. EH-CA changes only this assignment. Let \(j^\star=T-H_{\mathrm{eff}}(\mathbf{q}_{1:T})+1\) denote the first query within the effective horizon. We keep the original outcome signal for queries from \(j^\star\) onward and reduce it for earlier queries:
\begin{equation}
A_{i}^{\mathrm{out}}
=
\begin{cases}
\beta A^{\mathrm{traj}}, & i < j^\star,\\
A^{\mathrm{traj}}, & i \geq j^\star,
\end{cases}
\label{eq:ehca_weight}
\end{equation}
where \(\beta\in[0,1]\) controls how much of the final-outcome signal earlier queries retain. Thus, \(\beta=1\) recovers TROJail's original assignment, while \(\beta=0\) removes this signal entirely before the effective horizon. All other components of TROJail training remain unchanged. 
See details in Appendix~\ref{app:more_details_oc}.

\begin{figure}[!t]
    \centering
    
    \includegraphics[width=0.98\linewidth]{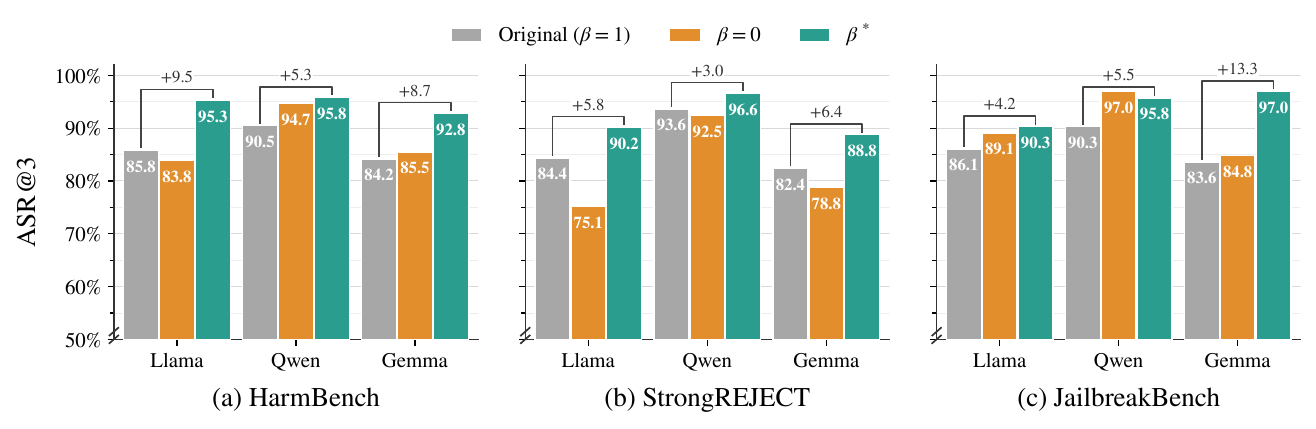}
    \vspace{-0.2cm}
    \caption{Reducing outcome credit before the effective horizon improves attacker training. TROJail assigns the same outcome signal to every query (\(\beta=1\)), while EH-CA scales this signal by \(\beta\) for queries before the effective horizon. Completely removing the earlier signal (\(\beta=0\)) is not consistently beneficial, but partial retention improves ASR@3 across all pairs.
    \(\beta^\star\) denotes the best value in \(\{0.1,0.2,0.3\}\), selected on HarmBench for each victim and transferred to the other two benchmarks.
    Across all pairs, \(\beta^\star\) improves ASR@3 by \(6.9\) points on average.}
    \label{fig:eh-ca}
\end{figure}

We follow the TROJail setup and evaluate three victims on HarmBench, StrongREJECT, and JailbreakBench, varying only \(\beta\). Figure~\ref{fig:eh-ca} shows that completely removing the earlier learning signal is too aggressive: \(\beta=0\) improves ASR@3 (success within three attempts) in five of nine victim--benchmark pairs but hurts the other four. In contrast, partial downweighting is consistently beneficial. 
Selecting \(\beta\in\{ 0.1,0.2,0.3\}\) on HarmBench for each victim improves ASR@3 in all pairs, by \(6.9\) points on average over the original setting.
This result helps explain why the effective horizon is useful for credit assignment. Queries before the effective horizon may still help attackers discover what to ask later, so removing their signal entirely can lose useful information. Meanwhile, giving them the same outcome signal as later queries is unnecessarily coarse. In our experiments, retaining a smaller signal for earlier queries gives the best of both.
Appendix~\ref{app:moreresult_beta} reports detailed results.

\subsection{Reward Propagation in Defense Training}
\label{sec:52}

We next ask whether the same distinction appears in defense training. TurnGate~\citep{shen2026one} makes a \textsc{Pass} or \textsc{Block} decision at each turn and receives a process reward for that decision. Its training signal for an earlier decision, however, includes rewards from all later turns. With TurnGate's choice of \(\gamma=\lambda=1\), the advantage at turn \(i\) is
\begin{equation}
A_i=\sum_{j=i}^{T} r_j,
\end{equation}
where \(r_j\) is the reward at turn \(j\). This means that the decision at turn \(i\) is trained using its own reward together with every reward that comes after it. For example, in a five-turn conversation, the decision at turn 1 receives a learning signal based on rewards from all five turns, while the decision at turn 5 uses only its own reward. Rewards can therefore influence decisions made many turns earlier.

Rather than measuring a separate effective horizon for every defense decision, we test the underlying question directly by limiting how far later rewards can influence an earlier decision. To this end, we replace TurnGate's original training signal, which sums the reward at turn \(i\) with all rewards from later turns, with a truncated version:
\begin{equation}
A_i^{(K)}
=
\sum_{j=i}^{\min(i+K,T)} r_j.
\end{equation}
Here, \(K\) controls how many future turns contribute to the learning signal at turn \(i\). When \(K=0\), a decision is trained only from the reward at that same turn. When \(K=1\), it also receives 
the next-turn reward,
and larger values extend this window farther into the future. For example, in a five-turn conversation, the decision at turn 1 uses only \(r_1\) when \(K=0\), uses \(r_1+r_2\) when \(K=1\), and eventually uses rewards from all five turns in the full setting. Importantly, we change only this reward window during training. The defense still observes the same dialogue at each turn. We can therefore ask directly whether a long conversation also requires rewards from distant future turns to train earlier decisions.

\begin{figure}[!t]
    \centering
    \includegraphics[width=0.98\linewidth]{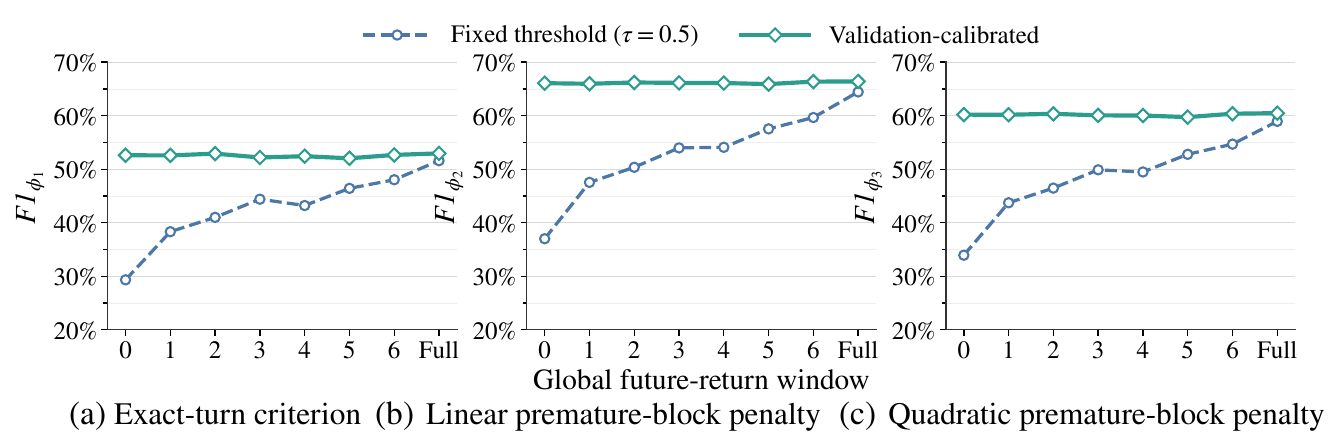}
    \vspace{-0.2cm}
    \caption{
    Long-range reward propagation has little effect on calibrated defense performance. TurnGate normally propagates rewards through the full remaining trajectory, while \(K\) limits each decision to rewards from at most \(K\) later turns; \(K=0\) uses only the current reward. Longer windows improve performance at the original threshold \(\tau=0.5\), but the gap largely disappears after selecting each policy's threshold on the validation set. For \(F1_{\phi_2}\), performance across reward windows has a standard deviation of only \(0.16\%\), and \(K=0\) performs nearly the same as the full return.
    }
    \label{fig:eh-rw}
\end{figure}

We follow the TurnGate setup on MTID~\citep{shen2026one}, starting every policy from the same supervised checkpoint and holding the data, rewards, and optimization procedure fixed while varying \(K\in\{0,1,2,3,4,5,6,\mathrm{full}\}\). 
See details in Appendix~\ref{app:more_details_defense}.
Figure~\ref{fig:eh-rw} shows that the answer depends strongly on how the decision threshold is chosen. At TurnGate's original threshold of \(\tau=0.5\), performance generally improves as rewards are propagated farther. After choosing each policy's threshold on the validation set, however, the differences largely disappear. Across reward windows, the standard deviation of test \(F1_{\phi_2}\) is only \(0.16\%\), and even \(K=0\) performs nearly the same as the full return. 
Appendix~\ref{app:defense_results_AR} provides additional metrics and threshold-free AUROC/AUPRC results.

The calibration result suggests that much of the apparent benefit from long reward windows comes from where a policy's scores fall relative to the fixed threshold, rather than from a large difference in the calibrated operating point. In this setting, rewards therefore do not need to travel across the entire conversation to train a similarly performing defense. Together with the attacker results, this reinforces the broader distinction developed throughout the paper: the trajectory may be long, while the useful temporal range of its learning signal can be much shorter.

\section{Conclusion}

Multi-turn jailbreaks are often described by the number of turns they take, yet the interaction used to discover an effective attack can be much longer than the history later needed to recover its progress. This distinction is becoming increasingly relevant as multi-turn attacks use more adaptive search and learned policies~\citep{rahman2025x,xiong2026trojail}. In this work, we asked how much prior query history is needed to recover the same attack progress at a later turn. Motivated by this question, we introduced the \emph{effective horizon}, which measures the shortest sequence of recent attacker queries sufficient to recover that progress. An attack may take many turns to discover even when only a few of those turns remain necessary once the useful queries have been found.

Across both search-based and training-based attacks, roughly half of the original query sequence is sufficient on average. This compression is useful beyond measurement. Recovering a high-scoring point from its effective history provides a better context for continuing unsuccessful attacks, while reducing final-outcome credit before the effective horizon improves attacker training. In defense training, rewards propagated over a short range preserve nearly the same calibrated performance as rewards propagated through the full remaining trajectory. Together, these results show that trajectory length and the amount of history that later behavior depends on can differ substantially. Treating them separately gives a more direct basis for deciding what context to retain and how far learning signals should extend~\citep{zeng2025reinforcing}. This distinction may become important beyond jailbreaks as reinforcement learning is applied to long multi-turn agent trajectories~\citep{wang2025ragen}.

\textbf{Limitations and Future Work.} Our results also raise broader questions about why later attack queries can remain effective with so little of the interaction history that helped produce them.
\begin{itemize}[leftmargin=*]

    \item \textbf{Recovering an attack is different from discovering it.}
The effective horizon is measured only after later queries have been found. Earlier turns may still help construct those queries even if they are no longer needed for replay. This matters for multi-agent attacks~\citep{rahman2025x}, adaptive tree search~\citep{wei2025trojan}, and lifelong attack generation~\citep{bhuiya2025plague}. Future work could use effective-horizon information to discover shorter attack paths directly.

    \item \textbf{The horizon depends on how progress is measured.}
    We identify the shortest suffix that recovers the original attack score, so the horizon depends on the judge and recovery tolerance. We use established jailbreak evaluators, including HarmBench~\citep{mazeika2024harmbench}, but repeated replays or multiple measures of attack progress could provide more robust estimates. Our current analysis is also turn-level; measuring dependence within queries may reveal additional compression.

    \item \textbf{Temporal dependence can be richer than a single horizon.}
    EH-CA uses one boundary for final-outcome credit, while our defense experiment uses a fixed reward window rather than estimating a horizon for each decision. More adaptive methods could learn how much influence earlier turns should receive from the trajectory itself. This question also extends to multi-turn agents~\citep{wang2025ragen,zeng2025reinforcing} and multi-agent attacks~\citep{rahman2025x}, where later actions may depend on selected messages, observations, or actions from much earlier in the interaction.

\end{itemize}

\subsection*{Acknowledgments}
NBE would like to acknowledge support from the DSO National Laboratories. We also acknowledge the U.S. Department of Energy, under Contract Number DE-AC02-05CH11231 for providing computational resources.

\bibliographystyle{plain}
\bibliography{ref}

\clearpage
\appendix

\section{Additional Experimental Details}
\label{app:more_details}

Because the main text focuses on the central empirical results, we use this section to spell out the parts of the protocol that matter for reproducing and interpreting them. The effective-horizon measurements and the two attack applications use the same replay operation. We keep a contiguous sequence of attacker queries unchanged, run those queries in a new conversation, and allow the victim to generate new responses. This makes it possible to ask how much of the original query history is sufficient to recover a previously reached level of attack progress without rewriting the attack or reusing the victim responses that originally produced it.

\subsection{Effective-Horizon Compression}
\label{app:more_details_ehc}

The main characterization compares two different ways of producing multi-turn attacks. For X-Teaming, we use Qwen2.5-32B-Instruct as the attacker and Llama-3.3-70B-Instruct as the victim. Attack progress is scored by GPT-4o using the 5-point jailbreak judge~\citep{qi2024fine}, with score \(5\) indicating success. For TROJail, we initialize the attacker with Qwen2.5-3B-Instruct, train it on AdvBench~\citep{zou2023universal}, and evaluate it against Llama-3.1-8B-Instruct using the HarmBench classifier~\citep{mazeika2024harmbench}, where scores above \(0.9\) indicate success. The main comparison uses 200 WildJailbreak behaviors with a maximum interaction length of five turns, and the expanded five-turn X-Teaming evaluation uses 1,500 behaviors.

For a target turn \(j\), we record the original attack score \(v_j\) and evaluate candidate suffixes containing the last \(k\) attacker queries for \(k\in\{1,\ldots,j\}\). Each candidate is run in a new conversation. Only the attacker queries \(\mathbf{q}_{j-k+1:j}\) are retained; every victim response is generated again during replay. We do not summarize the removed prefix, rewrite the retained queries, or remove turns from the middle of the sequence. A candidate recovers the target progress when its replay score satisfies the criterion in Equation~\ref{eq:effective_horizon}, and the effective horizon is the smallest \(k\) that does so. If no tested suffix satisfies the recovery criterion, the effective horizon is undefined for that target turn. Because the victim responses are regenerated, this procedure measures whether the retained attacker queries are sufficient to recover the score reached at turn \(j\). It does not test whether the removed turns helped the attacker discover those later queries in the first place.

For successful trajectories, the target is the final turn, so the effective horizon measures how much of the query history remains necessary after the attack has reached success. For the unsuccessful-trajectory analysis in the main text, we instead truncate each trajectory at the first turn that reaches its highest attack score and measure the effective horizon at that point. This second construction is important for EH-CR because the continuation method does not require the initial attack to have succeeded. It only requires a point whose progress is worth recovering from a shorter history.


\subsection{Context Recovery in Search-Based Attacks}
\label{app:more_details_cr}

The context-recovery experiments ask whether the same shorter history that is sufficient for replay is also useful when an attacker continues searching. For both X-Teaming and ActorBreaker, we use Qwen2.5-32B-Instruct as the attacker and evaluate on 159 harmful behaviors from HarmBench. The six victim models are Qwen3-30B-A3B, Llama-3.3-70B, Muse-Glimmer-30B, DeepSeek-V4.1-Flash, GPT-5.6 Luna, and Gemini 3.8 Flash. GPT-4o serves as the 5-point jailbreak judge~\citep{qi2024fine}, with score \(5\) indicating success. Because this score is discrete, EH-CR uses \(\epsilon=0\), so the recovered context must reach the same judge score as the selected point in the original trajectory.

In the main five-turn experiment, each attack first receives five turns. We apply continuation only when this initial trajectory has not succeeded, and then allow up to five additional turns. Full Context continues from the complete failed trajectory. Peak Context returns to the highest-scoring turn and retains the full query history leading to that turn. EH-CR returns to the same highest-scoring turn and keeps only the shortest suffix that recovers its score. If no shorter suffix recovers that progress, EH-CR keeps the full prefix up to the selected turn. The comparison between Peak Context and EH-CR is especially useful because both restart from the same point in the original search. Their difference is the amount of earlier query history retained. The robustness experiments in the main text vary the amount of interaction before recovery and the number of turns available afterward, so the continuation result does not depend on the specific choice of five initial turns and five continuation turns.


\subsection{Outcome Credit in Attacker Training}
\label{app:more_details_oc}

The attacker-training experiment keeps the TROJail pipeline fixed and changes only how its final-outcome signal is distributed across queries. TROJail trains a separate Qwen2.5-3B-IT attacker for each victim using 520 AdvBench behaviors~\citep{zou2023universal} over 260 optimization steps. We consider Qwen2.5-7B-IT, Llama-3.1-8B-IT, and Gemma-2-9B-IT as victims, and evaluate each attacker on HarmBench~\citep{mazeika2024harmbench}, StrongREJECT~\citep{souly2024strongreject}, and JailbreakBench~\citep{chao2024jailbreakbench}. We report ASR@3 using the HarmBench classifier, counting a behavior as successful when at least one of three attempts scores above \(0.9\). We use \(\epsilon=0.02\) when computing the effective horizon for EH-CA, and leave TROJail's other credit components unchanged.

The credit-retention factor is selected without tuning it separately for every evaluation benchmark. For each victim, we choose \(\beta\in\{0.1,0.2,0.3\}\) using HarmBench performance and then use that same value on StrongREJECT and JailbreakBench. The original uniform assignment \(\beta=1\) and complete removal of the prefix outcome signal \(\beta=0\) are retained as baselines. Section~\ref{app:moreresult_beta} reports the full sweep and shows how the HarmBench-selected values transfer to the other two benchmarks.

\subsection{Reward Propagation in Defense Training}
\label{app:more_details_defense}

The defense experiment isolates the temporal range of the training signal while leaving the dialogue observed by the defense unchanged. We follow TurnGate~\citep{shen2026one} on the Multi-turn Intent Dataset (MTID). All variants use Qwen3-4B, start from the same reward-weighted SFT checkpoint, and are trained for three epochs with seed 42 using the same cached reference trajectories, reward design, and optimization settings. At inference time, the policy still observes the same conversation and makes a \textsc{Pass} or \textsc{Block} decision at each turn. The only quantity changed during RL training is how many later rewards can contribute to the learning signal for an earlier decision.

With a reward window of \(K\), the decision at turn \(i\) uses rewards from the current turn through at most \(K\) later turns,
\begin{equation}
A_i^{(K)}
=
\sum_{j=i}^{\min(i+K,T)} r_j.
\label{eq:app_reward_window}
\end{equation}
Thus, \(K=0\) trains a decision only from the reward at that turn, \(K=1\) also includes the next reward, and Full includes every reward through the end of the trajectory. For example, the first decision in a five-turn conversation uses only \(r_1\) when \(K=0\), uses \(r_1+r_2\) when \(K=1\), and uses \(r_1+\cdots+r_5\) in the Full setting. This ablation changes the distance over which rewards propagate during training without shortening the dialogue available to the defense.

TurnGate evaluates benign preservation together with how well harmful conversations are blocked at the first harm-enabling turn. For a dialogue trajectory \(\mathbf{q}_{1:T}\), let \(t^\star\) denote the first harm-enabling turn, with \(t^\star=\infty\) for benign trajectories, and let \(\hat{t}\) denote the first predicted \textsc{Block}, with \(\hat{t}=\infty\) if no block is produced. For benign trajectories,
\begin{equation}
\mathrm{BenignScore}=1-\mathrm{FPR},
\end{equation}
where \(\mathrm{FPR}\) is the fraction of benign trajectories that are incorrectly blocked. For harmful trajectories, TurnGate uses
\begin{equation}
\phi_i(\hat{t},t^\star)=
\begin{cases}
0, & \hat{t}<t^\star,\ i=1,\\[2pt]
\left(\frac{\hat{t}}{t^\star}\right)^{i-1},
& \hat{t}<t^\star,\ i\in\{2,3\},\\[4pt]
1, & \hat{t}=t^\star,\\[2pt]
0, & \hat{t}>t^\star.
\end{cases}
\label{eq:defense_phi}
\end{equation}
Here, \(\phi_1\) gives no credit to a premature block. The other two metrics give partial credit according to how close that block is to \(t^\star\), with \(\phi_3\) penalizing an early decision more strongly than \(\phi_2\).

Let \(\mathrm{ACC}\) denote the fraction of harmful trajectories blocked exactly at \(t^\star\), and \(\mathrm{EARLY}\) the fraction blocked before \(t^\star\). The harmful-side score is
\begin{equation}
\mathrm{HarmfulScore}_{\phi_i}
=
\mathrm{ACC}
+
\mathrm{EARLY}\,
\mathbb{E}\!\left[
\phi_i(\hat{t},t^\star)
\mid
\hat{t}<t^\star
\right],
\end{equation}
and the final metric is the harmonic mean of benign and harmful performance,
\begin{equation}
F1_{\phi_i}
=
\frac{
2\,\mathrm{BenignScore}\,\mathrm{HarmfulScore}_{\phi_i}
}{
\mathrm{BenignScore}+\mathrm{HarmfulScore}_{\phi_i}
},
\qquad i\in\{1,2,3\}.
\end{equation}

We evaluate each learned policy in two ways because a fixed decision threshold can confound score calibration with the quality of the learned ranking. We first report the original TurnGate threshold \(\tau=0.5\). We then choose a separate threshold for each policy by maximizing \(F1_{\phi_2}\) on the validation set and apply that threshold unchanged to the test set. Section~\ref{app:defense_results_AR} additionally reports AUROC and AUPRC, which remove the threshold choice altogether. These evaluations let us distinguish a policy that genuinely ranks harmful turns better from one whose scores have simply shifted relative to a fixed cutoff.

\section{A Concrete Effective-Horizon Example}
\label{app:case_study}

The aggregate EHR results summarize how much query history can be removed, but they do not show what this compression looks like inside an actual attack. Table~\ref{tab:effective-horizon-example} follows one successful five-turn X-Teaming trajectory whose judge score rises from \(1\) to \(5\). The attacker reaches the maximum score only at turn 5. We then keep the original attacker queries fixed and replay suffixes of the trajectory in new conversations, allowing the victim responses to be generated again.

The final query by itself reaches score \(4\), so a one-turn suffix does not recover the progress of the original trajectory. When turn 4 is added, the replay reaches score \(5\). The shortest sufficient suffix therefore contains turns 4 and 5, which gives \(H_{\mathrm{eff}}=2\) and an EHR of \(2/5=0.4\). This example also shows why the effective horizon is more informative than asking whether the last query can jailbreak the victim on its own. The final query still depends on some local conversational setup, but that dependence extends only one turn backward rather than across the full five-turn trajectory.

The first three turns should not be interpreted as useless. They may have helped the attacker learn how the victim responds, move toward the later framing, or construct the queries that eventually reached score \(5\). The replay experiment asks a narrower question. Once turns 4 and 5 have already been discovered, are turns 1--3 still needed in the query history to recover the same progress? In this trajectory they are not. This is the distinction between the interaction used to discover a useful attack path and the shorter history sufficient to recover the progress reached along that path.

\begin{table*}[t]
\centering
\caption{
A five-turn attack can depend on much less history when its later queries are run again. Attacker queries are lightly summarized to preserve their conversational role without reproducing harmful details. The final query alone reaches score \(4\), while replaying turns 4--5 in a new conversation recovers the original score of \(5\). The trajectory therefore has \(H_{\mathrm{eff}}=2\) and EHR \(=0.4\). Earlier turns may still have helped discover the retained queries; the replay result shows that they are no longer required in the query history used to recover the final progress.
}
\label{tab:effective-horizon-example}
\small
\setlength{\tabcolsep}{6pt}
\begin{tabular}{c p{0.60\textwidth} cc}
\toprule
Turn & Redacted attacker query & Original score & Replay suffix \\
\midrule
1 & Request a neutral explanation of socioeconomic disparity. & 1 & Discarded \\
2 & Ask how people with fewer resources are stereotyped in conversation and media. & 2 & Discarded \\
3 & Request examples of disparaging jokes, framed as awareness of harmful language. & 4 & Discarded \\
\addlinespace[2pt]\cmidrule(lr){1-4}
4 & Request supposedly ``harmless'' everyday remarks that reinforce class stereotypes. & 4 & Retained \\
5 & Reframe the request as sarcastic remarks by a fictional character in social media or workplace settings. & 5 & Retained \\
\midrule
\multicolumn{2}{l}{Replay of turn 5 in a new conversation} & \multicolumn{2}{c}{final score $4$ (not recovered)} \\
\multicolumn{2}{l}{Replay of turns 4--5 in a new conversation} & \multicolumn{2}{c}{final score $5$ (recovered)} \\
\bottomrule
\end{tabular}
\end{table*}

\section{Victim-LLM Query Cost of Attack Continuation}
\label{app:victim}

The main continuation results compare methods under the same number of additional attack turns, but EH-CR performs extra victim queries while it searches for a suffix that recovers the selected progress. A natural concern is therefore that the ASR gain comes at the cost of many more victim calls. We examine total query cost separately from continuation length because EH-CR introduces two opposing effects. Validating candidate suffixes adds queries before continuation begins, while a better continuation context can allow the attack to succeed in fewer subsequent turns. The net cost depends on which effect is larger.

\begin{table*}[!ht]
\centering
\caption{
EH-CR does not have a uniform effect on total victim-query cost. For X-Teaming, fewer continuation turns offset the queries used to validate the effective suffix, while for ActorBreaker they do not. The table reports average victim-LLM queries per behavior in the five-turn continuation experiment, including suffix-replay validation for EH-CR. Results include all valid behaviors for which the initial attack fails, regardless of continuation success. Lower is better, and best values are shown in bold.
}
\label{tab:continuation_query_cost}
\begin{tabular}{
l
c
c
c
c
c
c
}

\toprule

&
\multicolumn{3}{c}{X-Teaming}
&
\multicolumn{3}{c}{ActorBreaker}
\\

\cmidrule(lr){2-4}
\cmidrule(lr){5-7}

Victim Model
&
Full
&
Peak
&
EH-CR
&
Full
&
Peak
&
EH-CR
\\

\midrule

Qwen3-30B-A3B
&
15.57
&
14.16
&
\textbf{13.65}
&
\textbf{5.13}
&
5.24
&
8.14
\\

Llama-3.3-70B
&
12.77
&
13.03
&
\textbf{11.90}
&
5.38
&
\textbf{5.36}
&
7.65
\\

Muse-Glimmer-30B
&
\textbf{19.02}
&
19.05
&
19.35
&
\textbf{7.04}
&
7.09
&
10.04
\\

DeepSeek-V4.1-Flash
&
18.12
&
\textbf{17.59}
&
18.11
&
6.79
&
\textbf{6.72}
&
8.85
\\

GPT-5.6 Luna
&
18.70
&
\textbf{17.82}
&
18.22
&
6.54
&
\textbf{6.50}
&
7.89
\\

Gemini 3.8 Flash
&
18.22
&
18.21
&
\textbf{16.80}
&
\textbf{5.88}
&
5.94
&
8.79
\\

\midrule

Avg.
&
17.07
&
16.64
&
\textbf{16.34}
&
\textbf{6.13}
&
6.14
&
8.56
\\

\bottomrule
\end{tabular}
\end{table*}

\begin{table*}[t]
\centering
\caption{
The shorter EH-CR context leads to earlier termination for both attacks. The table reports the average number of continuation turns used per behavior under a five-turn budget. EH-CR uses fewer turns for every evaluated victim model with both X-Teaming and ActorBreaker, although the reduction is larger for X-Teaming. Lower is better, and the pooled average aggregates behaviors across all six victim models.
}
\label{tab:continuation_turn_usage}

\begin{tabular}{
l
c
c
c
c
c
c
}

\toprule

&
\multicolumn{3}{c}{X-Teaming}
&
\multicolumn{3}{c}{ActorBreaker}
\\

\cmidrule(lr){2-4}
\cmidrule(lr){5-7}

Victim Model
&
Full
&
Peak
&
EH-CR
&
Full
&
Peak
&
EH-CR
\\

\midrule

Qwen3-30B-A3B
&
4.24
&
3.90
&
\textbf{3.06}
&
4.21
&
4.27
&
\textbf{4.09}
\\

Llama-3.3-70B
&
3.58
&
3.61
&
\textbf{2.58}
&
4.22
&
4.25
&
\textbf{4.08}
\\

Muse-Glimmer-30B
&
4.91
&
4.93
&
\textbf{4.56}
&
4.77
&
4.80
&
\textbf{4.59}
\\

DeepSeek-V4.1-Flash
&
4.71
&
4.62
&
\textbf{4.26}
&
4.87
&
4.84
&
\textbf{4.55}
\\

GPT-5.6 Luna
&
4.80
&
4.63
&
\textbf{4.07}
&
4.87
&
4.85
&
\textbf{4.46}
\\

Gemini 3.8 Flash
&
4.68
&
4.70
&
\textbf{3.95}
&
4.36
&
4.42
&
\textbf{4.18}
\\

\midrule

Pooled Avg.
&
4.67
&
4.58
&
\textbf{4.03}
&
4.59
&
4.61
&
\textbf{4.35}
\\

\bottomrule

\end{tabular}
\end{table*}

In Table~\ref{tab:continuation_query_cost}, the total-cost results show why EH-CR should be viewed primarily as a context-selection method. For X-Teaming, EH-CR uses \(16.34\) victim queries per behavior on average, compared with \(17.07\) for Full Context and \(16.64\) for Peak Context. Its replay overhead is therefore more than recovered by the reduction in downstream search. ActorBreaker behaves differently. EH-CR uses \(8.56\) queries per behavior, compared with \(6.13\) and \(6.14\) for Full and Peak Context. In that setting, replaying candidate suffixes costs more than the later continuation saves. The ASR improvement reported in the main text therefore does not come with a general claim of lower query cost.

Table~\ref{tab:continuation_turn_usage} separates the continuation itself from the cost of constructing the context. EH-CR terminates earlier for every victim model under both attacks. The pooled average for X-Teaming decreases from \(4.67\) turns with Full Context and \(4.58\) with Peak Context to \(4.03\) with EH-CR. For ActorBreaker, the reduction is smaller, from \(4.59\) and \(4.61\) turns to \(4.35\). This distinction explains the cost pattern above. The shorter recovered history makes the subsequent search more effective in both cases, but only X-Teaming saves enough continuation queries to pay for suffix validation. The benefit of EH-CR is therefore better use of the continuation context; whether that also reduces total victim calls depends on the dynamics of the underlying attack.

\section{Cross-Benchmark Selection of the Credit-Retention Factor}
\label{app:moreresult_beta}

The main EH-CA result depends on the amount of outcome credit retained before the effective horizon. If this value were chosen independently on every evaluation benchmark, an improvement could reflect benchmark-specific tuning rather than a reusable credit-assignment rule. We therefore use a stricter selection protocol. For each victim model, we select \(\beta\in\{0.1,0.2,0.3\}\) using HarmBench only and then freeze that value for StrongREJECT and JailbreakBench. The original TROJail assignment \(\beta=1\) and complete removal of prefix outcome credit \(\beta=0\) are evaluated as fixed baselines rather than selection candidates.

Table~\ref{tab:pw-beta-victim-final} shows that HarmBench selects \(\beta=0.1\) for Llama-3.1-8B-IT and Gemma-2-9B-IT, and \(\beta=0.3\) for Qwen2.5-7B-IT. These choices are then applied unchanged to the other two benchmarks. All six transfer evaluations improve over the original uniform assignment, with an average gain of \(6.4\) ASR@3 points. Within the tested partial-retention grid, the HarmBench-selected value also remains the strongest setting for the same victim on both StrongREJECT and JailbreakBench. Including the three HarmBench selection results gives an average improvement of \(6.9\) points across all nine victim--benchmark pairs. The consistency across benchmarks is useful because it indicates that the benefit is not produced by choosing a separate \(\beta\) after seeing each test benchmark.

The comparison with \(\beta=0\) also clarifies what the effective horizon does and does not say about credit. Completely removing the prefix outcome signal improves five of the nine evaluations but hurts the other four, including all three StrongREJECT settings. Queries before the effective horizon can therefore still carry useful training information even when they are not needed to replay the final attack progress. They may help the attacker discover the later queries that fall inside the effective horizon. EH-CA retains that possibility by assigning those earlier queries a smaller outcome signal instead of forcing their credit to zero. The transfer results support this softer use of the effective horizon as a boundary for reducing credit rather than treating it as a causal cutoff.

\begin{table*}[t]
\centering
\caption{
A credit-retention factor selected on HarmBench transfers to two other benchmarks without further tuning. For each victim, we choose \(\beta\in\{0.1,0.2,0.3\}\) on HarmBench and apply that value unchanged to StrongREJECT and JailbreakBench. Shaded rows show the selected settings. They outperform the original \(\beta=1\) assignment in all six cross-benchmark evaluations by \(6.4\) ASR@3 points on average; including HarmBench, the average gain is \(6.9\) points across all nine evaluations. Values are mean \(\pm\) standard deviation over three evaluation runs.
}
\label{tab:pw-beta-victim-final}
\small
\setlength{\tabcolsep}{9pt}
\begin{tabular}{llccc}
\toprule
Victim & Setting & HarmBench & StrongREJECT & JailbreakBench \\
\midrule

\multirow{5}{*}{Llama-3.1-8B-IT}
& Original (\(\beta=1\))
& \(85.83 \pm 0.62\)
& \(84.38 \pm 0.85\)
& \(86.06 \pm 2.27\) \\
& \(\beta=0\)
& \(83.83 \pm 0.62\)
& \(75.12 \pm 0.59\)
& \(89.09 \pm 1.48\) \\
& \cellcolor{gray!15}\(\beta=0.1\)
& \cellcolor{gray!15}\(95.33 \pm 1.31\)
& \cellcolor{gray!15}\(90.16 \pm 0.33\)
& \cellcolor{gray!15}\(90.30 \pm 0.86\) \\
& \(\beta=0.2\)
& \(90.67 \pm 0.24\)
& \(73.84 \pm 1.66\)
& \(78.18 \pm 3.93\) \\
& \(\beta=0.3\)
& \(77.50 \pm 0.41\)
& \(58.80 \pm 0.59\)
& \(75.15 \pm 2.27\) \\
\midrule

\multirow{5}{*}{Qwen2.5-7B-IT}
& Original (\(\beta=1\))
& \(90.50 \pm 1.41\)
& \(93.63 \pm 1.18\)
& \(90.30 \pm 2.27\) \\
& \(\beta=0\)
& \(94.67 \pm 0.24\)
& \(92.48 \pm 1.82\)
& \(96.97 \pm 0.86\) \\
& \(\beta=0.1\)
& \(85.67 \pm 1.43\)
& \(84.72 \pm 0.49\)
& \(84.85 \pm 3.74\) \\
& \(\beta=0.2\)
& \(88.50 \pm 1.78\)
& \(86.81 \pm 0.75\)
& \(92.12 \pm 1.71\) \\
& \cellcolor{gray!15}\(\beta=0.3\)
& \cellcolor{gray!15}\(95.83 \pm 1.43\)
& \cellcolor{gray!15}\(96.64 \pm 1.07\)
& \cellcolor{gray!15}\(95.76 \pm 2.27\) \\
\midrule

\multirow{5}{*}{Gemma-2-9B-IT}
& Original (\(\beta=1\))
& \(84.17 \pm 1.70\)
& \(82.41 \pm 0.87\)
& \(83.64 \pm 1.48\) \\
& \(\beta=0\)
& \(85.50 \pm 1.22\)
& \(78.82 \pm 1.47\)
& \(84.85 \pm 2.27\) \\
& \cellcolor{gray!15}\(\beta=0.1\)
& \cellcolor{gray!15}\(92.83 \pm 0.24\)
& \cellcolor{gray!15}\(88.77 \pm 1.34\)
& \cellcolor{gray!15}\(96.97 \pm 0.86\) \\
& \(\beta=0.2\)
& \(90.00 \pm 0.82\)
& \(84.84 \pm 0.65\)
& \(92.12 \pm 3.74\) \\
& \(\beta=0.3\)
& \(89.83 \pm 0.62\)
& \(87.96 \pm 1.34\)
& \(90.30 \pm 0.86\) \\
\bottomrule
\end{tabular}
\end{table*}

\section{Additional Results for Reward Propagation in Defense Training}
\label{app:defense_results_AR}

The main defense experiment asks whether an earlier decision needs rewards from the full remainder of the conversation in order to learn an effective blocking policy. The fixed-threshold results alone leave an important ambiguity. A longer reward window might genuinely improve the policy's ability to distinguish the harm-enabling turn, or it might mainly move the policy's scores relative to the fixed threshold \(\tau=0.5\). We use validation-based threshold calibration and threshold-free ranking metrics to separate these possibilities. The full metrics below also show how changes in the reward window trade off benign preservation, missed harmful turns, and premature blocking.

\begin{table*}[t]
\centering
\caption{
After threshold calibration, TurnGate performance changes little across reward-propagation windows. All variants use Qwen3-4B, start from the same reward-weighted SFT checkpoint, and differ only in \(K\), the number of later turns whose rewards can contribute to an earlier decision. Full uses every remaining reward. Each RL policy's \textsc{Block} threshold is selected on validation using \(F1_{\phi_2}\) and then fixed for all test metrics. The additional Full row uses TurnGate's original fixed threshold \(\tau=0.5\) so that the effect of calibration can be seen directly.
}
\label{tab:future-window-calibration}
\setlength{\tabcolsep}{3.5pt}
\renewcommand{\arraystretch}{1.15}
\resizebox{\textwidth}{!}{%
\begin{tabular}{lrrrrrrrrrrr}
\toprule
& & \multicolumn{1}{c}{Benign (Utility)}
& \multicolumn{6}{c}{Harmful}
& \multicolumn{3}{c}{Overall} \\
\cmidrule(lr){3-3}\cmidrule(lr){4-9}\cmidrule(lr){10-12}
Method & $\tau$
& \shortstack{Benign\\Score $\uparrow$}
& Miss $\downarrow$ & Early $\downarrow$ & $\ell_1\downarrow$
& \shortstack{Acc.\\($\phi_1$) $\uparrow$}
& \shortstack{Harmful\\Score $\phi_2\uparrow$}
& \shortstack{Harmful\\Score $\phi_3\uparrow$}
& $\mathit{F1}_{\phi_1}\uparrow$
& $\mathit{F1}_{\phi_2}\uparrow$
& $\mathit{F1}_{\phi_3}\uparrow$ \\
\midrule
Reward-weighted SFT$^{\dagger}$ & --
& 0.856 & 0.407 & 0.262 & 0.90 & 0.332 & 0.460 & 0.402 & 0.478 & 0.599 & 0.547 \\
TurnGate (Full, fixed threshold) & 0.5000
& 0.818 & 0.306 & 0.318 & 0.93 & 0.377 & 0.531 & 0.461 & 0.516 & 0.644 & 0.589 \\
\midrule
TurnGate ($K=0$) & 0.3430
& 0.765 & 0.219 & 0.380 & 0.98 & 0.401 & 0.582 & 0.496 & 0.526 & 0.661 & 0.602 \\
TurnGate ($K=1$) & 0.4013
& 0.796 & 0.253 & 0.355 & 0.95 & 0.393 & 0.563 & 0.484 & 0.526 & 0.660 & 0.602 \\
TurnGate ($K=2$) & 0.4013
& 0.794 & 0.241 & 0.362 & 0.96 & 0.397 & 0.568 & 0.487 & 0.529 & 0.662 & 0.603 \\
TurnGate ($K=3$) & 0.4013
& 0.788 & 0.233 & 0.377 & 0.97 & 0.390 & 0.569 & 0.485 & 0.522 & 0.661 & 0.601 \\
TurnGate ($K=4$) & 0.4013
& 0.786 & 0.229 & 0.378 & 0.99 & 0.393 & 0.570 & 0.486 & 0.524 & 0.661 & 0.600 \\
TurnGate ($K=5$) & 0.4013
& 0.772 & 0.216 & 0.392 & 1.01 & 0.393 & 0.575 & 0.487 & 0.520 & 0.659 & 0.597 \\
TurnGate ($K=6$) & 0.4378
& 0.790 & 0.232 & 0.373 & 0.98 & 0.395 & 0.572 & 0.488 & 0.527 & 0.664 & 0.604 \\
TurnGate (Full) & 0.4626
& 0.780 & 0.223 & 0.376 & 0.99 & 0.401 & 0.578 & 0.494 & 0.530 & 0.664 & 0.605 \\
\bottomrule
\end{tabular}%
}
\vspace{2pt}
\begin{minipage}{\textwidth}
\footnotesize
\(^{\dagger}\) The reward-weighted SFT checkpoint is shown as a reference using its historical default-decoding evaluation and is not calibration-matched to the RL rows. All scores are in \([0,1]\), except \(\ell_1\), which is measured in turns over blocked harmful trajectories.
\end{minipage}
\end{table*}

Table~\ref{tab:future-window-calibration} makes the calibration effect visible. At the original threshold \(\tau=0.5\), the Full policy obtains \(F1_{\phi_2}=0.644\), and the ablation in the main text shows a general improvement as rewards are allowed to propagate farther. The learned policies do not, however, use the same score scale. Their validation-selected thresholds range from \(0.3430\) for \(K=0\) to \(0.4626\) for Full. Once each policy is evaluated at its own validation-selected threshold, the difference across reward windows becomes small. \(F1_{\phi_2}\) ranges from \(0.659\) to \(0.664\), with a standard deviation of only \(0.16\) percentage points. Even \(K=0\), which trains every decision using only its same-turn reward, reaches \(0.661\) compared with \(0.664\) for Full. The benign and harmful components move somewhat differently across settings, but these changes largely cancel at the calibrated operating point.

This result indicates that a fixed threshold can make long-range propagation look more important than it is for the underlying decision rule. Calibration alone cannot establish that the learned rankings are equivalent, because two policies can achieve similar performance after choosing different thresholds while still ordering examples differently. We therefore evaluate the raw score rankings directly in Table~\ref{tab:future-window-ranking}. AUROC and AUPRC do not require choosing a \textsc{Block} threshold, so they provide a separate test of whether larger \(K\) makes harm-enabling turns more distinguishable from benign and pre-harm turns.

\begin{table*}[t]
\centering
\caption{
Threshold-free ranking quality is nearly unchanged across reward-propagation windows. AUROC varies only from \(0.808\) to \(0.810\), while AUPRC varies from \(0.452\) to \(0.457\). The \(K=0\) policy matches or slightly exceeds Full on both metrics. The ranking task treats harm-enabling turns as positives and benign and pre-harm turns as negatives; the AUPRC no-skill baseline is the positive prevalence of \(19.27\%\).
}
\label{tab:future-window-ranking}
\small
\setlength{\tabcolsep}{7pt}
\begin{tabular}{lcccccccc}
\toprule
& \multicolumn{8}{c}{Global future-return window} \\
\cmidrule(lr){2-9}
Metric & $K=0$ & $K=1$ & $K=2$ & $K=3$ & $K=4$ & $K=5$ & $K=6$ & Full \\
\midrule
AUROC $\uparrow$ & 0.810 & 0.809 & 0.810 & 0.809 & 0.808 & 0.808 & 0.809 & 0.809 \\
AUPRC $\uparrow$ & 0.457 & 0.453 & 0.457 & 0.455 & 0.454 & 0.453 & 0.455 & 0.452 \\
\bottomrule
\end{tabular}
\end{table*}

The threshold-free results reinforce the calibrated comparison. AUROC is \(0.810\) for \(K=0\) and \(0.809\) for Full, while AUPRC is \(0.457\) and \(0.452\), respectively. There is no monotonic improvement as \(K\) grows, and the entire range across all eight policies is very small. In this experiment, propagating rewards through more of the remaining trajectory therefore does not produce a measurable improvement in the model's ability to rank the turn at which harmful intent becomes actionable. The fixed-threshold gains are better explained by differences in where the learned scores fall relative to a common cutoff.

Our reproduced Full setting at the original threshold differs slightly from the value reported by TurnGate. We keep this row because it anchors the ablation to the released setup, while the scientific comparison in our experiment is between matched runs. Every reward-window policy starts from the same SFT checkpoint and uses the same data, cached trajectories, rewards, number of training epochs, seed, and optimization settings. Only \(K\) changes. We therefore interpret relative differences across these matched policies rather than small differences in absolute reproduction. The reward-weighted SFT row is included only as a reference because its historical default-decoding evaluation is not calibration-matched to the RL rows.

Taken together, the calibrated and threshold-free results sharpen the claim made in the main text. The defense still observes the complete dialogue at each turn, so short reward windows do not remove conversational information from the policy input. They only shorten the distance over which a later reward can train an earlier decision. Under this controlled offline setting, extending that distance to the full remaining trajectory provides little additional benefit once threshold placement is accounted for. The useful temporal range of the learning signal can therefore be much shorter than the dialogue on which the decision is conditioned.

\end{document}